# A Vision-Based Tactile Sensing System for Multimodal Contact Information Perception via Neural Network

Weiliang Xu, Guoyuan Zhou, Yuanzhi Zhou, *Student Member, IEEE*, Zhibin Zou, Jiali Wang, Wenfeng Wu, *Student Member, IEEE*, and Xinming Li, *Member, IEEE*

*Abstract*—Typically, robotic dexterous hands are equipped with various sensors to acquire multimodal tactile information, which is an important way for robots to perceive and interact with the environment. Vision-based tactile sensors have been widely developed due to their simple structure and high resolution. It adopts specialized optical design to convert different contact information into different signal responses, such as recognizing contact force information based on marking method, obtaining contact shape information based on photometric stereo method, and so on. However, the current mainstream vision-based tactile sensing systems tend to adopt isolated optical design strategies for different modal tactile information, introducing limitations in system integration. This article proposes a vision-based tactile sensing strategy for multimodal tactile information using only reflected light field information. The specific implementation of the vision-based tactile sensor and the recognition algorithm for sensing multiple tactile information simultaneously are described in detail. The results show that the tactile sensing strategy does not need to design different optical structures for different modal tactile information, but only uses a simple reflection layer combined with a neural network to sense multi-modal tactile information, which reduces the complexity of the tactile system. In addition, the system achieves a force error of 0.2 N and a pose error of 0.41 °, and shows excellent precision and recall in localization and classification tasks, demonstrating the potential for multimodal tactile integration in various fields.

## I. INTRODUCTION

THE dexterity of human manipulation relies on the thousands of sensory receptors in hands, which are used to detect various types of tactile information, including vibration, pressure, and friction [1-2]. This multimodal tactile information is also essential for robotic dexterous manipulation. To achieve accurate measurement of external physical stimuli, several tactile sensing mechanisms have been developed, such as piezoelectric [3-6], capacitive [7-10], and resistive [11-13] sensors. These tactile sensors exhibit a high-speed response, compact structure, and extensive range. However, most of these tactile sensors are based on point measurements, constrained by integration and signal acquisition issues, leading to low spatial resolution [14]. Additionally, due to the limitations of sensing principles, reliance on different sensing mechanisms for perceiving different tactile modes is necessary, making it challenging to achieve decoupled perception of multiple tactile modes within a single system [15]. This restricts their applications in compact sensor structures and multimodal tactile sensing tasks. Among different types of tactile sensors, vision-based tactile sensors have been widely studied due to their simple structure and high resolution [16-25]. These sensors utilize a deformable elastomer and a CMOS sensor to transform tactile information into visual representations. By modifying the optical design, vision-based tactile sensors capture different tactile information in the form of tactile images in various physical domains. Appropriate algorithms and models are then designed to achieve accurate feature extraction and recognition (such as force [16-17], shape [18-19], localization [14, 20], and contact posture [20-21]).

With the increasing diversity of tactile modalities, it becomes necessary to adopt different optical design strategies for different contact information. Reflective membrane-based method based on an elastomer surface with reflection pigment [21]. When an object makes contact with the elastomer surface, the deformation of surface reflects shadow images, making it suitable for precisely measuring the position and orientation of the contact object. Another approach using photometric stereo builds upon the reflective membrane-based method, utilizing a specially designed light source [22]. It calculates the direction gradient of the elastomer surface deformation based on the intensity of images captured from different light source directions. This method exhibits clear advantages in contact shape and texture perception. However, since contact geometry is a state information rather than a process information, it is difficult to measure the contact distribution force and other mechanical properties only with the contact geometry information [17]. Marker displacement-based method often involves preparing marker patterns on the elastomer surface. When the elastomer undergoes deformation

Manuscript summited Jan. , 2024. This work was supported by the Guangdong Basic and Applied Basic Research Foundation (No. 2022A1515010136), the Guangdong Provincial Key Laboratory of Nanophotonic Functional Materials and Devices, and the South China Normal University start-up fund. (Corresponding author: Xinming Li).

Weiliang Xu, Guoyuan Zhou, Yuanzhi Zhou, Zhibin Zou, Jiali Wang, Wenfeng Wu, and Xinming Li are with Guangdong Provincial Key Laboratory of Nanophotonic Functional Materials and Devices, School of Information and Optoelectronic Science and Engineering, South China Normal University, Guangzhou 510006, People's Republic of China. (e-mail: xmli@m.scnu.edu.cn)

due to contact, the feature points of the marker move accordingly, reflecting dynamic contact information such as contact force and slip. Compared to reflective membrane-based method, this method has lower spatial resolution. Although continuous marker patterns [23] and dense marker patterns [24] optimize the accuracy, resolution, and reliability, precise object recognition still poses certain challenges.

In order to simultaneously recognize multimodal tactile information, researchers have attempted to integrate various structural designs and data processing techniques. For example, marker displacement-based method has been combined with reflective membrane-based method [16], where images obtained from the reflection film are used to measure the position and orientation of objects, and marker displacement is utilized to estimate external contact forces. Photometric stereo-based method also tends to incorporate marker patterns to supplement dynamic contact information [24], enabling reconstruction of the contact surface and recognition of contact forces. Additionally, the combination of thermosensitive materials and marker displacement-based method allows for the temperature and force sensing [25]. While combining different methods can leverage the advantages of these diverse approaches to achieve multimodal tactile perceptions, the current mainstream optical design strategies for different modalities of tactile information are often isolated. This implies that vision-based tactile sensors face more complex challenges in system integration and data decoupling. Therefore, effectively obtaining and decoupling multimodal tactile information in tactile images for accurate feature extraction and recognition is a major challenge for vision-based tactile sensing systems.

To this end, this article proposes a multimodal tactile perception strategy for vision-based tactile sensors, which does not require markers or special light source designs. It solely utilizes a reflection layer design to extract surface deformation from tactile images for object category, posture, localization, and force estimation. The main contributions of this article are as follows: 1) We introduce a multimodal tactile perception strategy. We also elaborate on the design idea, specific implementation, and a neural network-based multimodal tactile extraction method. 2) We propose a set of tactile spatial resolution characterization methods based on imaging resolution targets, which can be used to characterize the tactile resolution of different vision-based tactile sensor designs. 3) We conducted validation experiments for different tactile mode recognitions. The experimental results demonstrate that the proposed strategy can simultaneously recognize various contact information with considerable accuracy in object classification, posture, localization, and force estimation.

The remainder of this paper is structured as follows: Section II presents the design principles and implementation of the multimodal tactile perception strategy for vision-based tactile sensors. Section III describes the neural network-based algorithm for extracting multimodal tactile information.

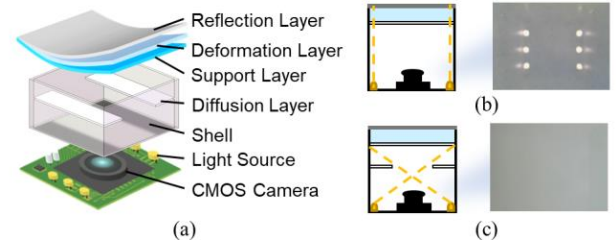

**Fig. 1.** (a) Schematic diagram of the vision-based tactile sensor. (b) The tactile image without diffusion layer design. (c) The tactile image with diffusion layer design.

Section IV experimentally evaluates the results. Section V provides a summary outlook.

## II. DESIGN OF THE VISION-BASED SENSOR

### A. Design of the Vision-Based Sensor

The structure of the proposed vision-based tactile sensor design is shown in **Fig. 1(a)**. The sensor consists of three main components: a soft contact interface with a reflection layer, a shell, and a CMOS image sensor (JX-H65) with light source. The soft contact interface consists of a reflection layer made of aluminum film manufactured by a sputter coater (Q150T S Plus, Quorum), a deformation layer made of polydimethylsiloxane (PDMS Sylgard 184, Dow), and a support layer made of acrylic, which converts the external physical contact information into optical information. The CMOS sensor equipped with a wide-angle lens is responsible for capturing optical information. The shell is 3D printed (S3 3D printer, Ultimaker) and has interfaces for installation. In the fabrication process, an acrylic sheet was fixed on a model made from the 3D printer to manufacture the support layer first. Then 2 mL of a mixture of uncured PDMS and curing agent at a ratio of 10:1 by mass was poured over the support layer to create a deformation layer, which was then placed in a vacuum chamber to remove air bubbles and allowed to cure naturally. Once the deformation layer was fully cured, the sample was plasma cleaned to improve the adhesion of coating. A reflection layer was then manufactured by sputtering an Al film onto the surface. Finally, the assembly of the CMOS sensor with LED module, the shell made of 3D printing, and the soft contact interface completed the overall design of the vision-based tactile sensor with an overall size of 35×35×30 mm. However, if all components were assembled directly, there would be overexposure on the tactile image due to the excessive light intensity at the light source projection on the reflection layer (**Fig. 1(b)**), thus introducing noise. Therefore, a shell with a diffusion layer was designed, as shown in **Fig. 1(c)**. When the light passes through the plane where the diffusion layer is located, it is attenuated before reaching the reflection layer, which eliminates the overexposure phenomenon.

### B. Mechanism of the Vision-Based Sensor

The working principle of vision-based tactile sensors is to capture the optical changes of the surface caused by contact.

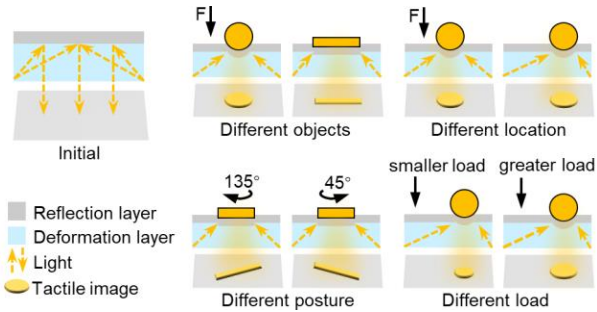

**Fig. 2.** Illustration of the principle of the tactile sensor.

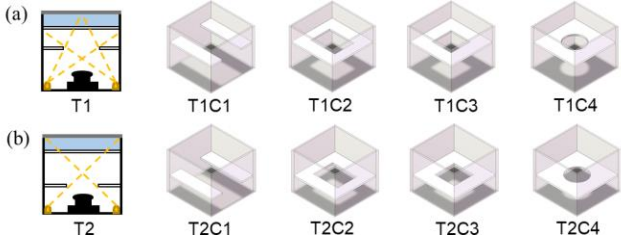

**Fig. 3.** (a) Four different diffusion pattern designs of Type 1. (b) Four different diffusion pattern designs of Type 2.

Here, assuming that the inner surface of the sensor is a diffuse reflecting surface, the surface can be described using a height function $z = f(x, y)$ [26]. In this case, the reflection intensity at the point $(x, y)$ can be represented as:

$$I(x,y) = R(\partial f/\partial x, \partial f/\partial y)E(x,y), \quad (1)$$

where $R$ is a reflective function defined based on the surface gradient $\partial f/\partial x$ and $\partial f/\partial y$, and $E$ represents the intensity of the incident light. When an object contacts the sensor, the surface of the sensor deforms, causing a change in the reflection function $R$ of the sensor surface. This change is characterized by local intensity variations in the image, allowing us to obtain tactile information about the deformation of the contact surface. As shown in **Fig. 2**, when an external force $F$ is applied to the flexible deformation layer, the optical reflectance under the reflection layer changes. The CMOS sensor at the bottom can sense the changes in the reflection field caused by different contact situations due to deformation. Information about different contact objects, localizations, and postures can be obtained by analyzing the distribution patterns of reflection field intensity in the tactile image. For different contact loads, the normal force can be estimated based on the reflection intensity and the area of the reflection field pattern. These details will be discussed in detail in the following sections.

*C. Design of Experiments for The Vision-Based Sensor*

As shown in **(1)**, the intensity of incident light is related to the decoupling of the final tactile information and the design of the diffusion layer reduces the intensity of incident light. By regulating the area size and the relative position of the diffusion layer, the light intensity on the reflection layer can be regulated to improve the system resolution of the sensor. Here, two different types of diffusion layers were designed by regulating their relative position. The first type is shown in **Fig. 3(a)** and is called Type 1 (T1). The diffusion layer is located at a position 18 mm above the CMOS sensor, allowing light to directly enter the surface of the reflection layer and illuminate it. The Type 2 (T2) is shown in **Fig. 3(b)**, where the diffusion layer is located at 10 mm above the CMOS sensor. The light enters the edge of the deformation layer, illuminating the inner surface of the reflection layer by a light-guided action of the deformation layer. In addition, with these two types, four different diffusion layer patterns were designed as shown in **Fig. 3(a)** and **Fig. 3(b)** by adjusting the light transmission area size. To ensure that the diffusion layer does not affect the CMOS imaging, with the camera lens diameter of 15 mm as the standard, a circular pattern with 15 mm diameter called Category 1 (C1), a square pattern with 15 mm side length called Category 2 (C2), a rectangular pattern with 15mm and 24mm side length called Category 3 (C3), and a rectangular with 15mm and 33mm side length called Category 4 (C4) were fabricated. It is worth mentioning that the regulations here are specific to the structure proposed in the previous section. In fact, the parameters of the diffusion layer are related to the sensor dimensions and the arrangement of light sources.

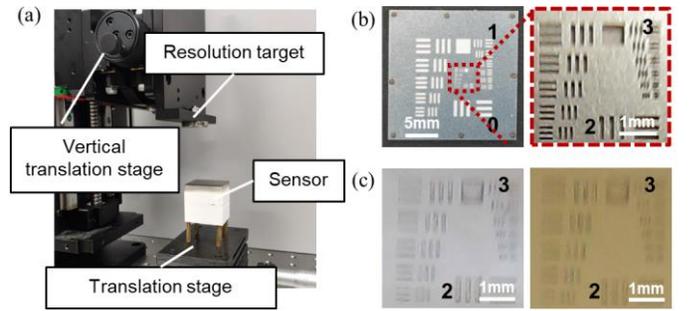

**Fig. 4.** Illustration of the principle of the tactile sensor. (a) The resolution test platform for the vison-based tactile sensor. (b) The resolution image of the CMOS camera. (c) The resolution image of the vision-based tactile sensor output using the T1C1 and T2C4.

To compare the resolution images of each design, we built a resolution test platform (**Fig. 4(a)**). Because vision-based tactile sensors use CMOS to capture contact information, the upper limit of tactile sensing resolution is theoretically the imaging resolution of a CMOS imaging system, which is 7.13 lp/mm for the horizontal stripe and 8.98 lp/mm for the vertical stripe. Therefore, a resolution test target based on the USAF 1951 resolution test target with 100 µm gradients was proposed (**Fig. 4(b)**). Using the proposed resolution test target as a quantitative standard enables a direct comparison of the resolution performance of different designs and their relationship to the theoretical upper image resolution limit.

The results are shown in **Fig. 5**, where **Fig. 5(a)** shows the comparison of the resolution performance of four different designs of T1 type and the original CMOS imaging system, and **Fig. 5(b)** shows the comparison of the resolution performance of four different designs of T2 type. It can be found that the average resolution of both horizontal and vertical stripes of the T1 type design is better than T2. The T1 design makes the light reaching the inner surface of the reflection layer more intense, thereby enhancing the resolution of the system to a certain extent. Similarly, for the same design type, when the area of the design pattern is larger, the intensity of the light reaching the inner surface of the reflection layer is more intense. Therefore, the C1 pattern design is the best. **Fig. 4(b) right** shows the resolution image of the camera, which represents the theoretical limit of spatial resolution for vision-based tactile sensors. **Fig. 4(c)** shows the tactile resolution images of TIC1 and T2C4, which have the best and the worst resolution performance, respectively. To conclude, the T1C1 design has a better performance with an average resolution of 7.13 lp/mm, which means that this design can resolve line pairs at the hundred-micron level comparable to the human tactile perception capability [27] and is the closest to the theatrical maximum tactile sensing resolution. Therefore, the T1C1 type will be used for the following experiments.

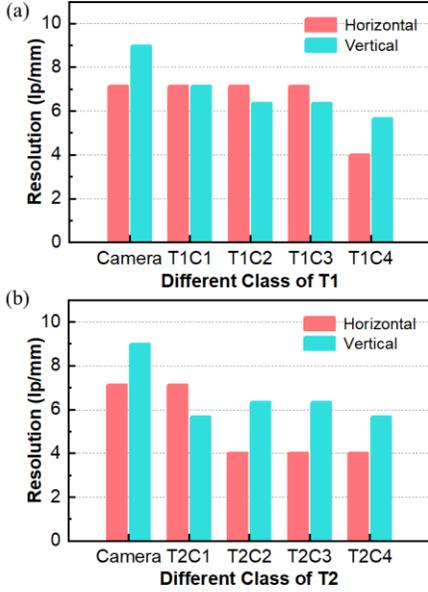

**Fig. 5.** (a) Resolution performance of four different designs of T1 and the original CMOS imaging system. (b) Resolution performance of four different designs of T2 and the original CMOS imaging system.

### III. DESIGN OF THE EXTRACTION ALGORITHM

#### A. Network Structure

When an object touches the vision-based tactile sensor, information about different contact objects, localizations, forces, and postures can be obtained by analyzing the distribution patterns of reflection field intensity in the tactile image. Here, a multimodal information extraction method based on neural network is introduced as shown in **Fig. 6(a)**, to accomplish contact objects classification, localization, postures angle, and normal force recognition. The structure of the network is shown in **Fig. 6(b)**. The original tactile images are passed through the backbone network to obtain 3 feature maps of different dimensions of size 80×80, 40×40, and 20×20. This can be seen by dividing the tactile image into a total of 8,400 regions. Then, the three feature maps of different dimensions are fused by FPN (Feature Pyramid Network) [28], and finally, the outputs of four different contact information in the 8400 regions are obtained after passing through a decoder with four channels (each channel has two 3 × 3 convolutional layers and one 1 × 1 convolutional layer). The first channel outputs the localization of the object in each region. The second channel outputs object category information, the third channel outputs posture angle information, and the fourth channel outputs normal force information. The PC used in training and testing the neural network has a GPU of NVIDIA GeForce RTX 3080, a CPU of Intel Xeon Gold 5222, and a system of Windows Server 2019. PyTorch framework, version 1.12.1, was utilized for model development.

#### B. Loss Function

The loss function can be divided into four blocks corresponding to four outputs. Firstly, the Binary Cross-Entropy loss ($BCE$) between the possibility of positive samples and the real situation of the 8400 regions is calculated. Then the normal force regression loss, the classification loss, the posture angle loss, and the loss of prediction location of the $B$ samples filtered by simOTA [29] are calculated. Here, to address the boundary problem in the pose angle recognition task, the angle values were encoded using the $CSL$ (Circular Smooth Label) technique [30] to convert the regression task into an angle classification task at a fine granularity of 1 °. Also, the $smoothL1$ (Smooth L1 Loss function) [31] was introduced to allow the network less sensitive to outliers and anomalies in the force regression task. Eventually, the loss function is expressed as follows,

$$Loss = \sum_{i=1}^{B} \Big\{ BCE(cls_{pre}, cls_{gt}) + BCE[\theta_{pre}, CSL(\theta_{gt})] \\ + smoothL1(f_{pre}, f_{gt}) \\ + \big[1 - IoU(box_{pre}, box_{gt})^2\big] \Big\} \\ + \sum_{i=1}^{8400} BCE(obj_{pre}, obj_{gt}), \quad (2)$$

where $B$ represents the total number of samples filtered by simOTA. 8400 represents the output dimension of the backbone. $obj_{pre}$ represents the probability that the prediction is a positive sample, and $obj_{gt}$ is 0 or 1 (where 0 represents a negative sample and 1 represents a positive sample). $cls_{pre}$ and $cls_{gt}$ represent the predicted and the real class, respectively. $\theta_{pre}$ and $\theta_{gt}$ represent the predicted and the real pose angle, respectively. $f_{pre}$ and $f_{gt}$ represent the predicted and the real normal force respectively. $CSL$ represents the $CSL$ encoding function with the Gaussian function as the window function. $smoothL1$ is a robust $L1$ loss that is less sensitive to outliers [30]. $IoU$ (Intersection over Union) evaluates the overlap of the predicted box $box_{pre}$ and the true box $box_{gt}$.

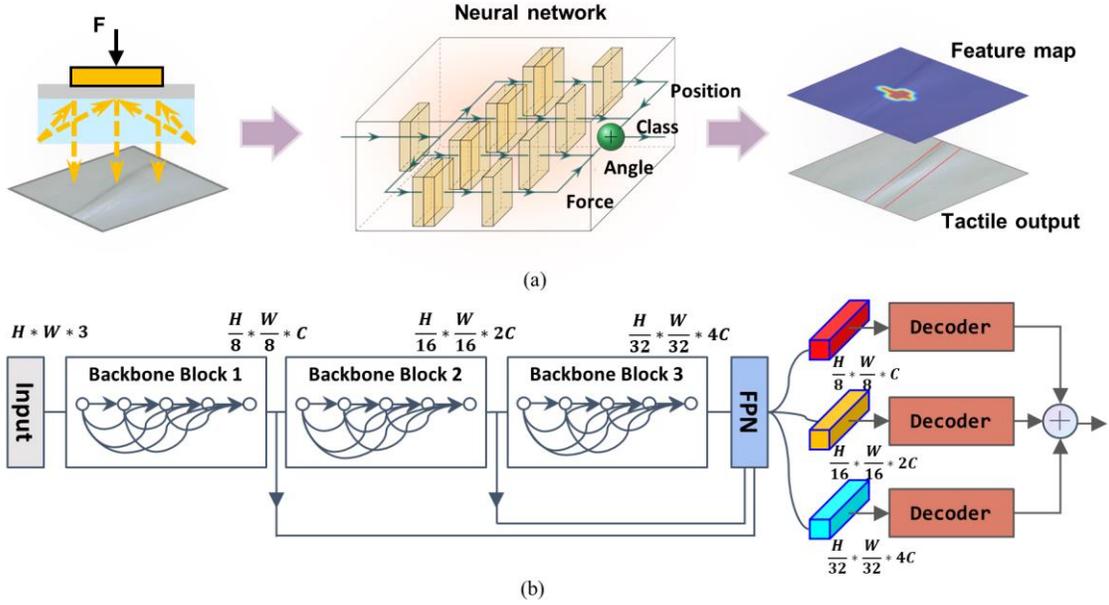

**Fig. 6.** (a) Architecture of the proposed data processing process. (b) The structure of the proposed network.

*C. Definition of the Pose Angle Error*

In the posture angle prediction task, the predicted angles range from 0° to 180°, and the angle numerical difference does not represent the actual angle difference, such as the difference between 1° and 179° is 178° in numerical value while the actual difference is only 2°. In this regard, as mentioned above, the $CSL$ technique was used to solve this problem during the training period. In the evaluation phase, we use the definition of rotation matrix to get the error of the pose angle shown as follows,

$$\Delta\theta = \arccos\left[\left|\frac{tr(R_{pre}R_{gt}^{-1})}{2}\right|\right], \quad (3)$$

where $R_{gt}$ represents the rotation matrix of the ground truth angle. $R_{pre}$ represents the rotation matrix of the predicted angle. $tr$ represents the trace of the matrix. $R_{gt}^{-1} \cdot R_{pre}$ can characterize the rotation matrix of the predicted value angle to the true value angle, and the derivation process is as follows.

As shown in **Fig. 7**, we define the actual and predicted angle as $\theta_1$ and $\theta_2$, and the actual frame $A_1$ and predicted frame $A_2$ of the target can be regarded as rotated by $\theta_1$ and $\theta_2$ along the x-axis $A_0$. So, $A_1$ and $A_2$ can be expressed as follows,

$$A_1 = \begin{bmatrix} x_1 \\ y_1 \end{bmatrix} = \begin{bmatrix} cos\theta_1 & -sin\theta_1 \\ sin\theta_1 & cos\theta_1 \end{bmatrix}\begin{bmatrix} x_0 \\ y_0 \end{bmatrix} = C_0^1 A_0, \quad (4)$$

$$A_2 = \begin{bmatrix} x_2 \\ y_2 \end{bmatrix} = \begin{bmatrix} cos\theta_2 & -sin\theta_2 \\ sin\theta_2 & cos\theta_2 \end{bmatrix}\begin{bmatrix} x_0 \\ y_0 \end{bmatrix} = C_0^2 A_0. \quad (5)$$

Here, $C_0^1$ represents $R_{gt}$ and $C_0^2$ represents $R_{pre}$. $A_2$ can be also considered as $A_1$ rotated by $\Delta\theta$, which represents the error of the posture angle. So $A_2$ can also be expressed as follows,

$$A_2 = \begin{bmatrix} x_2 \\ y_2 \end{bmatrix} = \begin{bmatrix} cos\Delta\theta & -sin\Delta\theta \\ sin\Delta\theta & cos\Delta\theta \end{bmatrix}\begin{bmatrix} x_1 \\ y_1 \end{bmatrix} = C_1^2 A_1. \quad (6)$$

By associating **(4)**, **(5)**, and **(6)**, we can find that the rotation matrix $C_1^2$ from $A_1$ to $A_2$ can be expressed as,

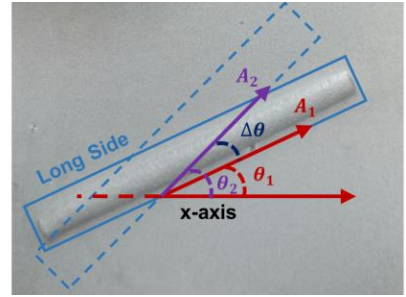

**Fig. 7.** Diagram of angle error calculation.

$$C_1^2 = C_0^2 C_1^0 = \begin{bmatrix} cos\Delta\theta & -sin\Delta\theta \\ sin\Delta\theta & cos\Delta\theta \end{bmatrix}. \quad (7)$$

Therefore, the absolute error between the actual angle and the predicted angle is finally obtained as **(3)**.

IV. EXPERIMENTAL EVALUATION

For accurate robot manipulation tasks, it is necessary to obtain multimodal contact information such as class, posture, location, and force applied to the target object during contact, etc. In this regard, this section demonstrates the performance of the proposed system in recognizing multimodal contact information through experiments. The experimental setup, as shown in **Fig. 8**, consists of a digital force gauge, a Raspberry Pi, and a translation platform. The sensor is fixed on the translation platform to control the contact location and pose angle. A digital force gauge is mounted on the vertical translation stages which is controlled to establish contact between the probe and the sensor. To investigate the relationship between tactile images and normal forces, we collected more than 3,000 samples of five spherical contact probes with diameters of 10 mm, 15 mm, 20 mm, 25 mm, and 30 mm. For posture angle analysis, we used a strip-shaped contact probe for collecting different posture data with more than 2,000 samples. In addition, we developed a contact probe with a contact interface that aligns with the proposed vision-based tactile sensor, allowing us to simulate scenarios where a robot grasps a screw and when it grasps five different objects.

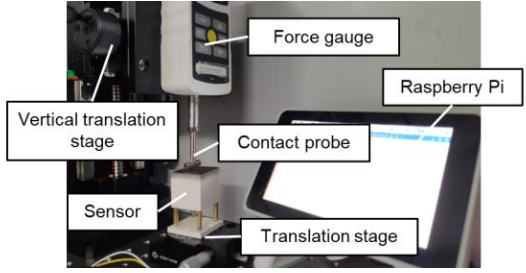

**Fig. 8.** The experimental setup of multimodal contact information recognition.

*A. Visual Characteristics of Different Contact Situations*

The vision-based tactile sensor converts tactile information into high-density, high-resolution tactile images, which can be decoupled as multimodal tactile feature information from redundant visual information. To investigate the ability of the multimodal vision-based tactile sensing system to distinguish between different modes of tactile information, we used different types of contact probes as test objects and analyzed their visual characteristics separately.

First, an induction system based on single-point contact was used to investigate the recognition of normal force. Five spherical contact probes of varying diameters were used to apply different loads. Because the spherical probes are geometrically isotropic, there is no issue of pose recognition during contact. By using probes of different diameters, the relationship between force conditions and tactile image features can be explored intuitively. As shown in **Fig. 9(a)**, three contact probes were selected to demonstrate that the normal force applied to the target object is related to the area and light field distribution of the tactile image. Under the same contact force, the tactile image produced by a larger diameter contact probe has a larger contact area and weaker contrast. For example, in the load range of 0 to 3 N, as the diameter of the contact probe increases, the tactile image has a larger contact area, and the contrast of the edge contrast gradually decreases. When the same diameter probe was used for contact, the tactile image produced under higher load conditions had a larger contact area and stronger contrast. For instance, when using a 10 mm diameter contact probe for contact, as the load increases, the resulting tactile image has a larger contact area, the intensity of the edge reflection light field gradually decreases, and the contrast of its edge contrast gradually increases. Additionally, because the radius of the contact area is correlated with the probe radius and load [32]. When using a smaller force with a larger-diameter contact probe and a larger force with a smaller-diameter contact probe, there may not be a significant difference in the contact area of the tactile images, as shown in **Fig. 9(a)**, images of (0~3 N, 30 mm), (3~6 N, 20 mm) and (6~10 N, 10 mm). However, in this scenario, the smaller-diameter contact probe creates a deeper indentation, causing the pixels corresponding to the center of the contact area to become brighter, resulting in a more pronounced edge contrast. Conversely, when using a larger-diameter contact probe with a smaller force, the indentation depth is shallower, causing the pixels at the center of the contact area to be relatively darker, and the edge contrast is comparatively weaker.

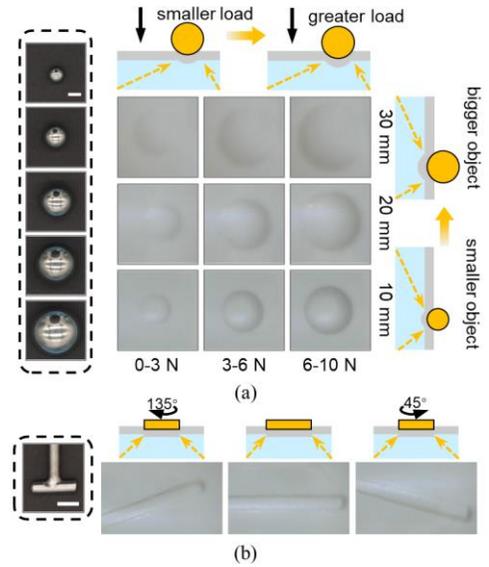

**Fig. 9.** Tactile images of different contact situations (the scale in the figure is 10 mm). (a) Tactile images of different normal forces. (b) Tactile images of different contact postures.

Furthermore, the proposed vision-based tactile sensing system can characterize different contact poses. A strip probe (**Fig. 9(b)**) was introduced to apply different contact poses to the sensor. For ease of subsequent data processing and discussion, a five-parameter method [31], with an angle range of 0° to 180°, was used to define posture information. For different contact postures, the tactile image presents different posture angles, and the distribution of the reflection light field is related to the object's contact posture angle. Additionally, as shown in the comparison between **Fig. 9(a)** and **Fig. 9(b)**, the proposed sensing system can also characterize different contact positions and object information, such as shape and type.

*B. Integrating Neural Network Enables Tactile Sensing for Different Modes.*

To validate the accuracy of the system in recognizing multimodal tactile through neural network system, we collected more than 3,000 samples with varying loads and more than 2,000 samples with different pose. Where all the samples were divided into train and test sets in a ratio of nine to one and 10 % of the samples in the train set were taken as the validation set. Furthermore, we compare three different backbone networks, a lightweight backbone ShuffleNet [33], a common backbone for target detection CSPnet [34], and a transformer-based image processing backbone Swin Transformer [35]. All three backbone networks compress the original image by a factor of 8, 16, and 32, so they can obtain feature maps with three different dimensions of 80×80, 40×40,

TABLE I
COMPARISON OF THREE DIFFERENT BACKBONES

| Backbone | MAE force (N) | MAE angle (°) | MAE location (mm) | Precision | FPS |
|---|---|---|---|---|---|
| CSPnet | **0.2** | **0.41** | **0.15** | 98.65 % | 28 |
| ShuffleNet | 0.7 | 0.67 | 0.21 | 98.65 % | **31.2** |
| Swim Transformer | 1.1 | 0.45 | 0.17 | **98.88 %** | 19 |

and 20×20 for decoder input.

**Table I** presents the evaluation results of the network for posture, normal force, localization, and classification. The results indicate that the performance of the three backbone networks is close in terms of localization recognition and classification tasks. For posture angle recognition, CSPnet backbones shows relatively better performance. Anyway, the model's discriminative limit is to distinguish a 1° angular difference, therefore the average accuracy of the three backbones falls within the systematic error range. In terms of normal force recognition, CSPnet outperforms the other two backbone networks significantly. The Frames per Second (FPS) is related to the number of network parameters, and the model with ShuffleNet has the best FPS. Theoretically, the more model parameters, the better the recognition accuracy, so the network with CSPnet as the backbone performs better overall than ShuffleNet. On the other hand, the Swin Transformer, with the largest number of parameters, lacks some of the inductive biases inherent to convolutional neural networks, leading to poor generalization in this study despite having more parameters. In summary, the model with CSPnet demonstrates better performance on the dataset in this experiment, and therefore, CSPNet is chosen as the backbone for subsequent experiments.

The detailed performance of the model is shown in **Fig. 10**. **Fig. 10(a)** shows the box plot of the normal force prediction error within the range of 0 to 10 N. The absolute prediction errors in all force ranges are within 0.3 N, and the MAE for all samples is 0.2 N. Specifically, when we narrow down the dataset range to the 0-3 N interval, the system's MAE can reach 0.1 N, with the error distribution shown in **Fig. 10(b)**. **Fig. 10(c)** shows the box plot of the estimation error in different angular ranges. The average prediction error in different pose angles is within 1°, and only a few samples have a prediction error of up to 2°. The experimental results show that, except for the system error introduced in manual data labeling, the proposed system has relatively high accuracy in pose recognition tasks. The system's error distribution for localization is shown in **Fig 10(d)**, which has a localization MAE of 0.15 mm, showing stable localization accuracy across the entire surface.

The experimental results show that distinct tactile features associated with different contact situations were discernible in the corresponding contact images. In contrast, conventional approaches to multimodal tactile detection often entail the integration of structures based on disparate mechanisms, thereby introducing additional complexity to hardware design and data processing. In light of these observations, our findings demonstrate that the proposed sensing system can effectively exploit image-based features to represent diverse tactile information and achieve high-performance recognition through the integration of machine learning techniques. Thus, our results underscore the potential advantages of utilizing visual representations of tactile information in the context of multimodal tactile sensing.

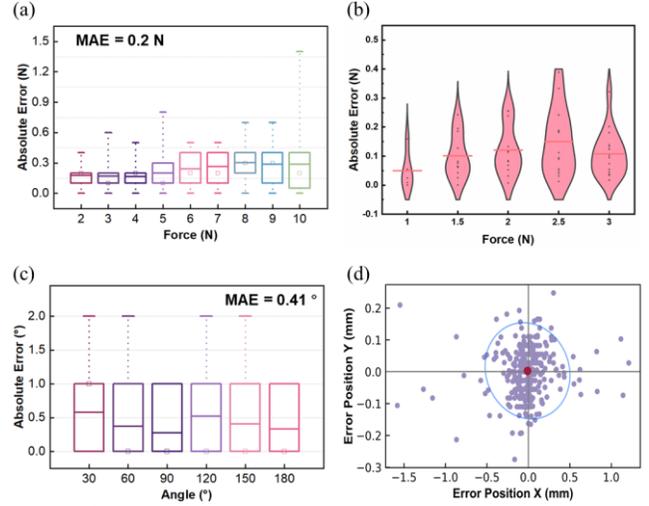

**Fig. 10.** (a) Box plot of absolute error of normal force identification within 0~10 N. (b) Violin plots of absolute error of normal force identification within 0~3 N. (c) Box plot of the absolute error of posture identification. (d) Error ellipse for localization errors.

*C. Vision-Based Tactile Sensing Intuitively and Accurately Identified Multimodal Tactile Information*

Multimodal tactile information, including object posture, category, localization, and force, plays a crucial role in the precise manipulation of objects. For instance, during grasping, multimodal tactile feedback can provide the grasping force and position feedback, ensuring the accurate grasping of objects. The proposed sensing system in the previous section enables us to quantitatively perceive the multimodal tactile information of objects during the grasping process by decoupling tactile images. Here, as shown in **Fig. 11**, we simulated four grasping scenarios of a screw (including the head, body, top, and bottom contact points) and collected more than 2,000 samples dividing the train set and test set in a ratio of 9 to 1. The various performances of the proposed sensing system in the grasping simulation for a screw are shown in **Fig. 12**. The error of normal force and pose angle for different contact parts are shown in **Fig. 12(a)**. The MAE of normal force identification for contacting the head, body, top, and bottom are 0.19 N, 0.21 N, 0.15 N, and 0.17 N respectively, and the posture angle identification errors for different contact parts are also all within 1 °. In addition, **Fig. 12(b)** and **(c)** respectively show the results of localization and contact part classification in the screw grasping scenarios, both achieving excellent outcomes.

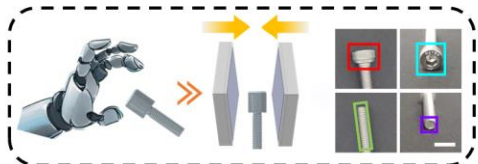

**Fig. 11.** Schematic diagram of the grasping simulation for four cases (the scale in the figure is 10 mm).

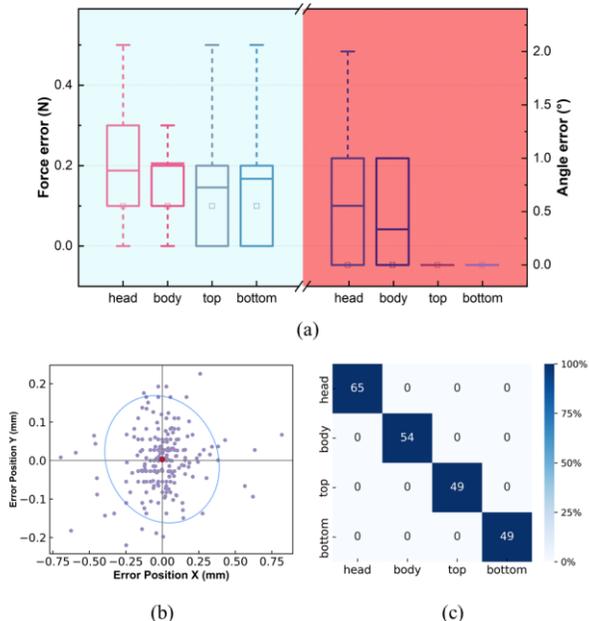

**Fig. 12.** (a) Box line plot of the estimation error of normal force and pose angle for different contact parts. (b) Error ellipse for localization errors. (c) Confusion matrix diagram for contact part classification.

In addition, to demonstrate the universality and practicality of distinguishing different tactile modes, as shown in **Fig. 13(a)** and **(b)**, we collected more than 2,000 samples of five common objects (a USB plug, a LEGO brick, an amplifier IC, a potentiometer, and a screwdriver), which then were divided as the train set and test set in a ratio of 9 to 1. Here, the AP@50 (Average Precision with 50 % as the threshold) value [36], that is, the harmonic means of recall and precision when positive samples are considered for the predicted and ground truth samples with $IoU$ greater than 50 %, was chosen as the metric to evaluate the estimation error of contact object location and contact part classes. As shown in **Fig. 14(a)**, the AP values of the amplifier IC, LEGO brick, and potentiometer all reach 100 %. The AP values of the screwdriver and USB plug are 94.81 % and 97.06 %, respectively, where the recall and precision of the screwdriver are 89.66 % and 92.86 %, and that of the USB plug is 97.14 % and 94.44 %. The violin plots of normal force and pose angle errors for different objects are shown in **Fig. 14(b)**. The mean absolute errors of normal force recognition for the USB plug, LEGO brick, amplifier IC, potentiometer, and screwdriver are 0.26 N, 0.17 N, 0.18 N, 0.17 N, and 0.27 N, respectively. And the overall MAE of the normal force for the test set is 0.2 N. The mean absolute errors of the pose angle recognition of the five different objects are all within 1 °, and the overall MAE of the pose angle for the test set is 0.61 °. In addition, there is no significant difference between the error distributions of normal force recognition and pose angle recognition for the five types of objects.

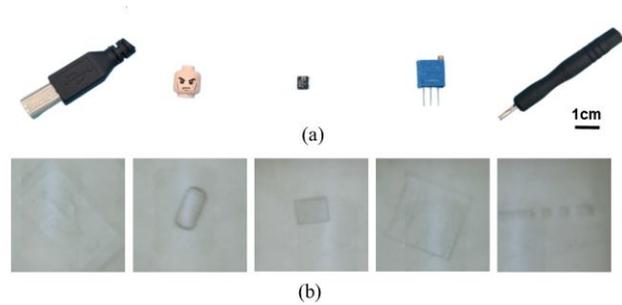

**Fig. 13.** (a) Schematic diagram of five objects. From left to right are a USB plug, a Lego brick, an amplifier IC, a potentiometer, and a screwdriver. (b) Tactile images of five objects.

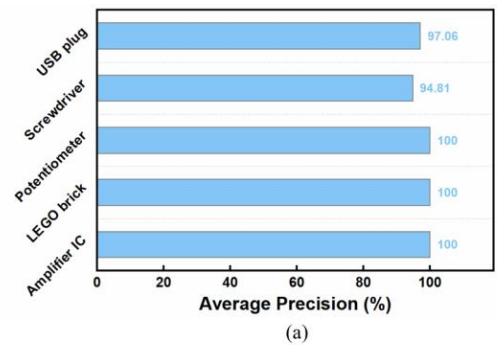

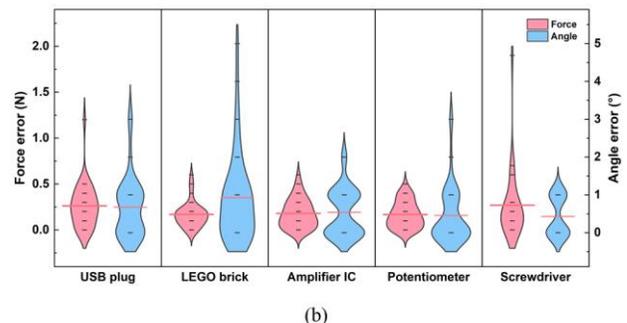

**Fig. 14.** (a) AP values for five common objects. (b) Violin plots of errors in normal force recognition and pose angle recognition for five objects.

The system achieves impressive performance in the classification and localization of contact objects, while maintaining a decent capability in normal force recognition and posture angle recognition, which is maintained under the same level for different contact cases as in the above experiments. Overall, the results illustrate that the proposed system can perform a perception of contact information for different modalities using only one sensor and without different data processing methods for different contact information while maintaining decent performance in each task. This ability is critical for a wide range of applications that require accurate manipulation of objects, such as manufacturing, robotics, and prosthetics. A video demonstration of the system's performance is also available (see **movie S1**).

## V. CONCLUSION

Accurately obtaining multimodal tactile information has been a challenge for most tactile sensing systems. This article proposed a multimodal tactile perception strategy for vision-based tactile sensors, and a specific implementation plan is provided. Furthermore, a set of tactile spatial resolution characterization methods based on imaging resolution targets and a multimodal tactile extraction method based on deep neural networks are introduced. Compared to traditional methods, the proposed strategy does not require markers or special light source designs. It solely utilizes a reflection layer design to extract surface deformation from tactile images, enabling the estimation of object category, posture, localization, and force. Experimental results demonstrate that the proposed strategy achieves a contact posture recognition error of 0.41°, a localization accuracy of 0.15 mm, force accuracy within the range of 0-10 N at 0.2 N, and accuracy within the range of 0-3 N at 0.1 N. Moreover, it can easily and accurately recognize various types of contact objects with excellent precision and recall, demonstrating good tactile feature recognition capabilities. The proposed strategy's reliance on the deformation of elastomer may lead to adverse effects due to the creep effect, especially in applications such as high-speed measurements under dynamic contact. In general, we believe that the strategy proposed in this article provides a new solution for multimodal tactile perception, offering high resolution, high integration, and effective tactile information decoupling. It is expected to be applicable to various fields requiring multimodal tactile sensing tasks, including biomedical, biology, and robotics. Additionally, since the tactile extraction method of this strategy is data-driven, and the extracted elastomer deformation features are generic for other tasks, we anticipate that in the future, it can be further expanded to recognize more tactile modalities information by simply increasing the output dimensions without the need for a redesign of data processing methods.

## REFERENCES

[1] A. Damasio and G. B. Carvalho, 'The nature of feelings: evolutionary and neurobiological origins', Nat Rev Neurosci, vol. 14, no. 2, Art. no. 2, Feb. 2013.
[2] J. Kim, B.-J. You, and Y. Choi, 'D'Arsonval Movement-Based Precise Milligram Force Control for Individual Touch Sense Assessment', IEEE Transactions on Industrial Electronics, vol. 64, no. 2, pp. 1534–1543, Feb. 2017.
[3] H. Meng, L. Zhou, X. Qian, and G. Bao, 'Design and Application of Flexible Resistive Tactile Sensor Based on Short-Circuit Effect', IEEE Transactions on Instrumentation and Measurement, vol. 72, pp. 1–8, 2023.
[4] W. Lin, B. Wang, G. Peng, Y. Shan, H. Hu, and Z. Yang, 'Skin-Inspired Piezoelectric Tactile Sensor Array with Crosstalk-Free Row+Column Electrodes for Spatiotemporally Distinguishing Diverse Stimuli', Advanced Science, vol. 8, no. 3, p. 2002817, 2021.
[5] S. Xie, Y. Zhang, H. Zhang, and M. Jin, 'Development of Triaxis Electromagnetic Tactile Sensor with Adjustable Sensitivity and Measurement Range for Robot Manipulation', IEEE Transactions on Instrumentation and Measurement, vol. 71, pp. 1–9, 2022.
[6] J. Wang et al., 'Energy-efficient, fully flexible, high-performance tactile sensor based on piezotronic effect: Piezoelectric signal amplified with organic field-effect transistors', Nano Energy, vol. 76, p. 105050, 2020.
[7] P. Maiolino, M. Maggiali, G. Cannata, G. Metta, and L. Natale, 'A flexible and robust large scale capacitive tactile system for robots', IEEE Sensors Journal, vol. 13, no. 10, pp. 3910–3917, 2013.
[8] M. Alshawabkeh et al., 'Highly Stretchable Additively Manufactured Capacitive Proximity and Tactile Sensors for Soft Robotic Systems', IEEE Transactions on Instrumentation and Measurement, vol. 72, pp. 1–10, 2023.
[9] Y. Xie, L. Lin, L. Lu, Q. Liang, and B. Zhang, 'Flexible Interfacial Capacitive Pressure Sensor Based on Randomly Distributed Micro-Pits Electrode', IEEE Transactions on Instrumentation and Measurement, vol. 71, pp. 1–9, 2022.
[10] C. Tang, X. Chang, J. Wang, Y. Peng, H. Wu, and H. Wang, 'A Non-Array Soft Capacitive Tactile Sensor with Simultaneous Contact Force and Location Measurement for Intelligent Robotic Grippers', IEEE Transactions on Instrumentation and Measurement, 2023.
[11] H. Zhu, H. Luo, M. Cai, and J. Song, 'A Multifunctional Flexible Tactile Sensor Based on Resistive Effect for Simultaneous Sensing of Pressure and Temperature', Advanced Science, p. 2307693, Dec. 2023.
[12] S. Pyo, J. Lee, K. Bae, S. Sim, and J. Kim, 'Recent Progress in Flexible Tactile Sensors for Human-Interactive Systems: From Sensors to Advanced Applications', Advanced Materials, vol. 33, no. 47, p. 2005902, Nov. 2021.
[13] H. Meng, L. Zhou, X. Qian, and G. Bao, 'Design and Application of Flexible Resistive Tactile Sensor Based on Short-Circuit Effect', IEEE Transactions on Instrumentation and Measurement, vol. 72, pp. 1–8, 2022.
[14] S. Cui, R. Wang, J. Hu, J. Wei, S. Wang, and Z. Lou, 'In-hand object localization using a novel high-resolution visuotactile sensor', IEEE Transactions on Industrial Electronics, vol. 69, no. 6, pp. 6015–6025, 2021.
[15] K. Park, H. Yuk, M. Yang, J. Cho, H. Lee, and J. Kim, 'A biomimetic elastomeric robot skin using electrical impedance and acoustic tomography for tactile sensing', Sci. Robot., vol. 7, no. 67, p. eabm7187, Jun. 2022.
[16] B. Fang et al., 'A dual-modal vision-based tactile sensor for robotic hand grasping', in 2018 IEEE International Conference on Robotics and Automation (ICRA), IEEE, 2018, pp. 4740–4745.
[17] B. Ward-Cherrier et al., 'The TacTip Family: Soft Optical Tactile Sensors with 3D-Printed Biomimetic Morphologies', Soft Robotics, vol. 5, no. 2, pp. 216–227, Apr. 2018.
[18] E. Donlon, S. Dong, M. Liu, J. Li, E. Adelson, and A. Rodriguez, 'Gelslim: A high-resolution, compact, robust, and calibrated tactile-sensing finger', in 2018 IEEE/RSJ International Conference on Intelligent Robots and Systems (IROS), IEEE, 2018, pp. 1927–1934.
[19] D. F. Gomes, Z. Lin, and S. Luo, 'GelTip: A finger-shaped optical tactile sensor for robotic manipulation', in 2020 IEEE/RSJ International Conference on Intelligent Robots and Systems (IROS), IEEE, 2020, pp. 9903–9909.
[20] S. Q. Liu, L. Z. Yañez, and E. H. Adelson, 'GelSight EndoFlex: A Soft Endoskeleton Hand with Continuous High-Resolution Tactile Sensing', in 2023 IEEE International Conference on Soft Robotics (RoboSoft), IEEE, 2023, pp. 1–6.
[21] K. Nozu and K. Shimonomura, 'Robotic bolt insertion and tightening based on in-hand object localization and force sensing', in 2018 IEEE/ASME International Conference on Advanced Intelligent Mechatronics (AIM), IEEE, 2018, pp. 310–315.
[22] W. Yuan, S. Dong, and E. H. Adelson, 'Gelsight: High-resolution robot tactile sensors for estimating geometry and force', Sensors, vol. 17, no. 12, p. 2762, 2017.
[23] M. Li, L. Zhang, T. Li, and Y. Jiang, 'Continuous marker patterns for representing contact information in vision-based tactile sensor: Principle, algorithm, and verification', IEEE Transactions on Instrumentation and Measurement, vol. 71, pp. 1–12, 2022.
[24] W. K. Do, B. Jurewicz, and M. Kennedy, 'DenseTact 2.0: Optical tactile sensor for shape and force reconstruction', in 2023 IEEE International Conference on Robotics and Automation (ICRA), IEEE, 2023, pp. 12549–12555.
[25] X.-X. Shi, Y. Chen, H.-L. Jiang, D.-L. Yu, and X.-L. Guo, 'High-Density Force and Temperature Sensing Skin Using Micropillar Array with Image Sensor', Advanced Intelligent Systems, vol. 3, no. 8, p. 2000280, Aug. 2021.
[26] K. Shimonomura, 'Tactile image sensors employing camera: A review', Sensors, vol. 19, no. 18, p. 3933, 2019.

[27] J. Dargahi and S. Najarian, 'Human tactile perception as a standard for artificial tactile sensing—a review', Int. J. Med. Robotics Comput. Assist. Surg., vol. 1, no. 1, pp. 23–35, Jun. 2004.
[28] T.-Y. Lin, P. Dollár, R. Girshick, K. He, B. Hariharan, and S. Belongie, 'Feature pyramid networks for object detection', in Proceedings of the IEEE conference on computer vision and pattern recognition, 2017, pp. 2117–2125.
[29] Z. Ge, S. Liu, F. Wang, Z. Li, and J. Sun, 'YOLOX: Exceeding YOLO Series in 2021'. arXiv, Aug. 05, 2021.
[30] X. Yang and J. Yan, 'On the Arbitrary-Oriented Object Detection: Classification Based Approaches Revisited', Int J Comput Vis, vol. 130, no. 5, pp. 1340–1365, May 2022.
[31] R. Girshick, 'Fast r-cnn', in Proceedings of the IEEE international conference on computer vision, 2015, pp. 1440–1448.
[32] M. Machado, P. Moreira, P. Flores, and H. M. Lankarani, 'Compliant contact force models in multibody dynamics: Evolution of the Hertz contact theory', Mechanism and machine theory, vol. 53, pp. 99–121, 2012.
[33] X. Zhang, X. Zhou, M. Lin, and J. Sun, 'Shufflenet: An extremely efficient convolutional neural network for mobile devices', in Proceedings of the IEEE conference on computer vision and pattern recognition, 2018, pp. 6848–6856.
[34] C.-Y. Wang, H.-Y. M. Liao, Y.-H. Wu, P.-Y. Chen, J.-W. Hsieh, and I.-H. Yeh, 'CSPNet: A new backbone that can enhance learning capability of CNN', in Proceedings of the IEEE/CVF conference on computer vision and pattern recognition workshops, 2020, pp. 390–391.
[35] Z. Liu et al., 'Swin transformer: Hierarchical vision transformer using shifted windows', in Proceedings of the IEEE/CVF international conference on computer vision, 2021, pp. 10012–10022.
[36] M. Everingham, S. M. A. Eslami, L. Van Gool, C. K. I. Williams, J. Winn, and A. Zisserman, 'The Pascal Visual Object Classes Challenge: A Retrospective', Int J Comput Vis, vol. 111, no. 1, pp. 98–136, Jan. 2015.

interests include Human-Computer Interaction and intelligent system design.

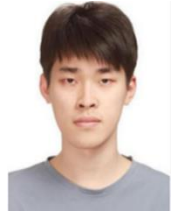
**Zhibin Zou.** He has received the B.S. degree from Huaqiao University, Xiamen, China. He is currently pursuing his M.S. degree in the School of Information and Optoelectronic Science and Engineering from South China Normal University, China. His current interests include optoelectronic technology.

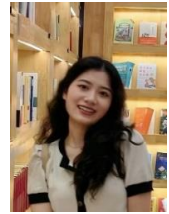
**Jiali Wang** received her bachelor's degree at the School of Physics and Electronic Engineering, Sichuan Normal University. She is currently studying for a master's degree in South China Normal University. Her current research interests are flexible sensors based on optics.

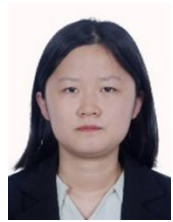
**Wenfeng Wu.** She received her bachelor's degree from the School of Electronics and Information Technology, Sun Yat-sen University. She is currently pursuing a master's degree in the School of Information and Optoelectronic Science and Engineering at South China Normal University. Her current interests include intelligent sensors.

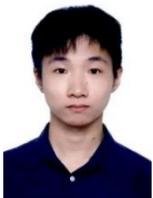
**Weiliang Xu** received B.S. degree in South China Normal University, Guangzhou, China, in 2022. He is currently pursuing his Ph.D. in the School of Information and Optoelectronic Science and Engineering at the South China Normal University. His current interests include Human-Computer Interaction and optoelectronic intelligent sensing design.

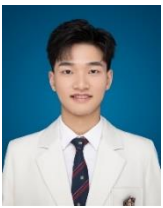
**Guoyuan Zhou** received B.S. degree in Optical Information Science Engineering from Nanjing Tech University. He is currently pursuing Ph.D. in South China Normal University. His current interests include vision-based tactile sensors.

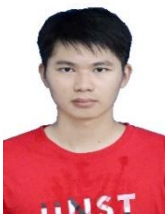
**Yuanzhi Zhou.** He received B.S. degree in the School of Physics, Hefei University of Technology, Hefei, China. He is currently pursuing his Ph.D. in South China Normal University, Guangzhou, China. His current

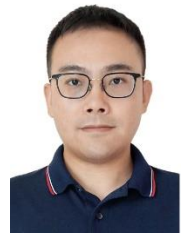
**Xinming Li** received his B.Sc. and Ph.D. degrees from Tsinghua University in 2007 and 2013, respectively. After that, he was an Assistant Professor at the National Center for Nanoscience and Technology, China, postdoctoral research fellow at The Chinese University of Hong Kong, research fellow at The Hong Kong Polytechnic University, and JSPS postdoctoral Fellow at the National Institute for Materials Science, Japan. He is currently a Professor with the School of Information and Optoelectronic Science and Engineering, at South China Normal University, China. He has co-authored more than 90 journal publications. His research interests include wearable devices, optoelectronic technology, vision-based tactile sensors, intelligent sensing systems, and haptics technology. He served as editor for Sensors, and Frontiers in Physics.